\documentclass[runningheads]{llncs}
\usepackage[T1]{fontenc}
\usepackage{graphicx,verbatim}
\usepackage{adjustbox}
\usepackage{multirow}
\usepackage{pifont}
\usepackage{amssymb} 
\usepackage{amsmath}
\usepackage{cleveref}
\usepackage{CJKutf8}
\usepackage{xcolor}
\usepackage[ruled,linesnumbered]{algorithm2e}

\usepackage{algpseudocode} 
\usepackage{lineno}
\usepackage[normalem]{ulem}
\usepackage{makecell}

\definecolor{r}{RGB}{255,0,0}
\definecolor{g}{RGB}{0,255,0}
\definecolor{b}{RGB}{0,0,255}
\usepackage{environ}
\NewEnviron{hide}{}

\newif\ifshowchanges
\showchangestrue   

\begin{document}
%
\title{Prompt Group-Aware Training for Robust Text-Guided Nuclei Segmentation}
%

\author{Yonghuang Wu, Zhenyang Liang, Wenwen Zeng, Xuan Xie, Jinhua Yu}  
\authorrunning{Anonymized Author et al.}
\institute{Fudan University, Shanghai, China \\
    \email{yonghuangwu21@m.fudan.edu.cn}}
  
\maketitle              
\begin{abstract}
Foundation models such as Segment Anything Model 3 (SAM3) enable flexible text-guided medical image segmentation, yet their predictions remain highly sensitive to prompt formulation. Even semantically equivalent descriptions can yield inconsistent masks, limiting reliability in clinical and pathology workflows.

We reformulate prompt sensitivity as a group-wise consistency problem. Semantically related prompts are organized into \emph{prompt groups} sharing the same ground-truth mask, and a prompt group-aware training framework is introduced for robust text-guided nuclei segmentation. The approach combines (i) a quality-guided group regularization that leverages segmentation loss as an implicit ranking signal, and (ii) a logit-level consistency constraint with a stop-gradient strategy to align predictions within each group. The method requires no architectural modification and leaves inference unchanged.

Extensive experiments on multi-dataset nuclei benchmarks show consistent gains under textual prompting and markedly reduced performance variance across prompt quality levels. On six zero-shot cross-dataset tasks, our method improves Dice by an average of 2.16 points. These results demonstrate improved robustness and generalization for vision-language segmentation in computational pathology.

\end{abstract}

\section{Introduction}

Foundation models have reshaped image segmentation by introducing promptable and highly generalizable architectures.
Segment Anything (SAM)~\cite{kirillov2023sam} enables a single model to support diverse prompts, achieving strong zero-shot performance across domains.
This flexibility has motivated growing interest in applying SAM-like models to medical and pathology imaging.

However, recent studies reveal that prompt-driven segmentation models are highly sensitive to prompt formulation, particularly in medical settings~\cite{mazurowski2023sammedical}.
Semantically equivalent text prompts may produce inconsistent segmentation outcomes.
In pathology images, prompts such as ``nuclei'', ``all cell nuclei'', or implicit subtype descriptions often refer to the same target but lead to unstable predictions, undermining reliability for clinical deployment.

Text-guided and open-vocabulary segmentation methods further explore language-conditioned dense prediction.
CLIP-based approaches~\cite{luddecke2022clipseg,li2022lseg,ding2022maskclip} and medical vision-language models~\cite{wang2022medclip} align visual and textual representations to enable segmentation from free-form descriptions.
Despite their success, these methods typically assume a one-to-one correspondence between a prompt and its target region, and performance remains sensitive to prompt phrasing.
Prompt variability is rarely modeled explicitly during supervised training.

Related robustness studies address annotation uncertainty, noisy user interactions, or adversarial prompts~\cite{zhang2023learning,li2024prism}.
Yet ambiguity is generally treated as noise to be mitigated, rather than as structured equivalence among multiple valid prompts describing the same segmentation target.
In pathology, linguistic variability is intrinsic: different expressions may legitimately correspond to identical anatomical or cellular structures, naturally inducing a many-to-one mapping from prompts to masks.

\begin{figure}[t]
    \centering
    \includegraphics[width=\linewidth]{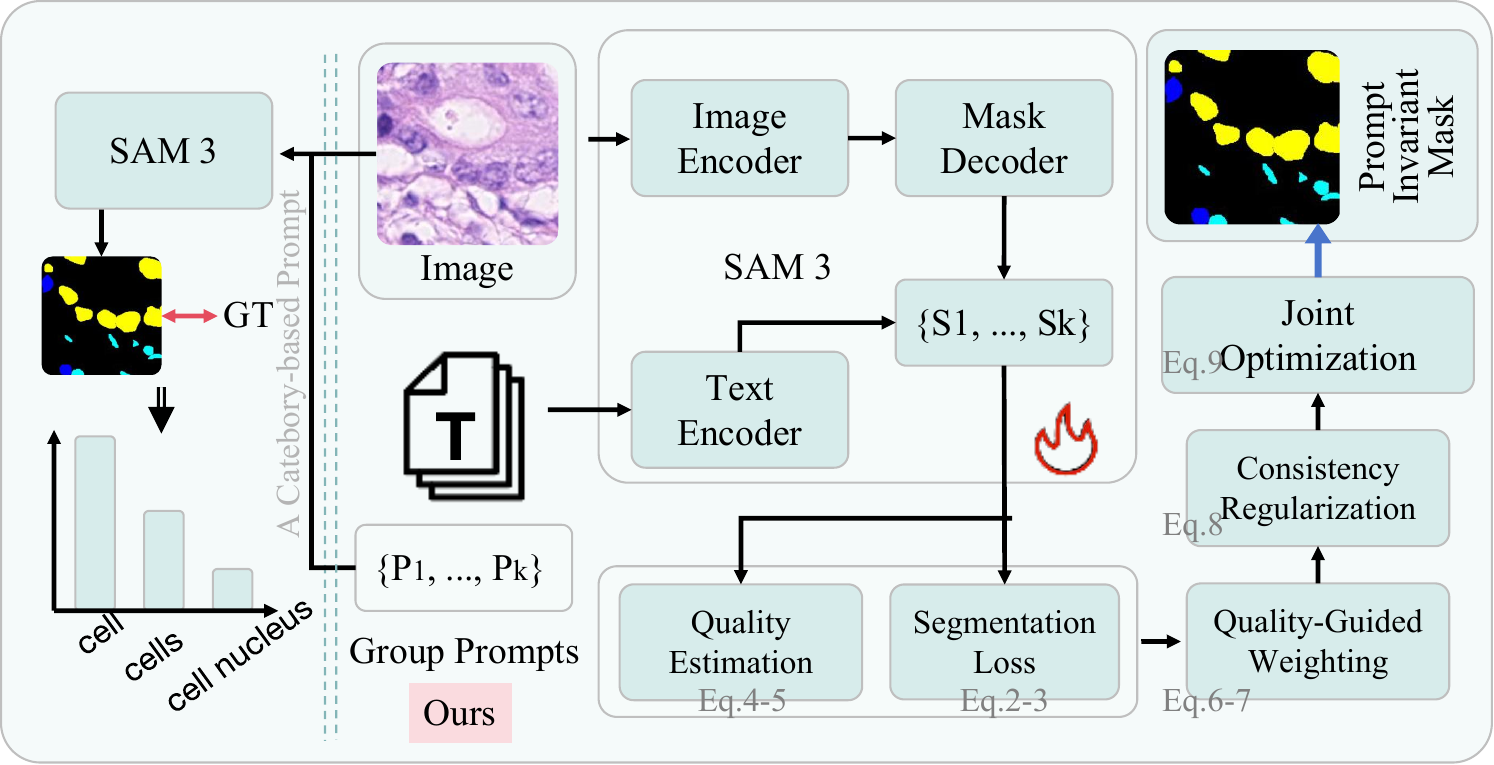}
    \caption{Overview of the proposed prompt group-aware training framework.}
    \label{fig:method}
\end{figure}

To address this gap, we reformulate prompt sensitivity as a group-wise consistency problem (see \ref{fig:method}).
For each image, we organize semantically related prompts into \emph{prompt groups} that share a common ground-truth mask.
We propose a group-aware training framework that enforces prompt-invariant behavior through two mechanisms:
(i) a quality-guided objective that models relative prompt reliability within each group, and
(ii) a prompt consistency loss that aligns segmentation outcomes across prompts.
Our method operates purely at training time, requires no additional supervision, and leaves the inference procedure unchanged.

Experiments on text-guided nuclei segmentation demonstrate that our approach improves accuracy while substantially reducing performance variance across diverse prompts, providing a practical pathway toward robust and trustworthy vision-language models in pathology.

\section{Method}

\subsection{Task Definition and Prompt Grouping}

We study referring image segmentation in pathology images under varying prompt quality.
Given an image $I \in \mathbb{R}^{H \times W \times C}$ and a set of natural language prompts
$\mathcal{P} = \{p_1, \dots, p_K\}$ that refer to the same target structure, the goal is to predict a prompt-invariant binary segmentation mask.
A text-conditioned segmentation model produces one mask per prompt:
\begin{equation}
S_i = f_\theta(I, p_i), \quad S_i \in \{0,1\}^{H \times W}.
\end{equation}

In pathology, semantically equivalent prompts may differ substantially in clarity or specificity (e.g., ``nuclei'' vs. implicit subtype descriptions), making text a potentially noisy conditioning signal.
To explicitly model prompt equivalence, we organize training data into \emph{prompt groups}.
Each group is defined as $(I, \mathcal{P}_g, M_g)$, where all prompts $\mathcal{P}_g$ correspond to the same ground-truth mask $M_g$, inducing a many-to-one mapping from prompts to supervision.

Prompt groups naturally include (i) \textbf{all-instance groups}, targeting all structures of interest, and (ii) \textbf{class-specific groups}, focusing on a particular cell type.
This formulation enables prompt-invariant learning by sharing supervision across diverse linguistic expressions, while mitigating conflicting gradients from semantically equivalent prompts.

\subsection{Group-wise Segmentation and Prompt Quality Modeling}

For a given image $I$ and prompt group $\mathcal{P}_g$, the model produces one prediction per prompt:
\begin{equation}
S_i = f_\theta(I, p_i).
\end{equation}

Each prediction is optimized with the standard SAM-style segmentation loss:
\begin{equation}
\mathcal{L}_{seg}^{(i)} = 
\mathcal{L}_{mask}^{(i)}
+ \mathcal{L}_{dice}^{(i)}
+ \mathcal{L}_{presence}^{(i)},
\end{equation}
The image encoder is shared across prompts, while the text encoder and mask decoder are applied independently, preserving the original architecture and training protocol.

\paragraph{Prompt Quality Estimation}

Prompts in the same group may vary in specificity. We quantify prompt quality using the segmentation loss:
\begin{equation}
q_i = - \mathcal{L}_{seg}^{(i)},
\end{equation}
where $ \mathcal{L}_{seg}^{(i)} $ is computed using the same top-$K$ selection and max-aggregation strategy as in inference.

We consider only relative quality within each prompt group:
\begin{equation}
\tilde{q}_i = q_i - \frac{1}{K}\sum_{j=1}^K q_j.
\end{equation}
All quality scores are detached from the computational graph and serve only as a relative ranking signal within each group.

\paragraph{Quality-Guided Group-aware Loss}

We further define a soft weighting scheme to modulate the contribution of different prompts:
\begin{equation}
w_i =
\frac{
\exp(-\mathcal{L}_{seg}^{(i)} / \tau)
}{
\sum_{j=1}^{K} \exp(-\mathcal{L}_{seg}^{(j)} / \tau)
},
\end{equation}
where $ \tau $ is a temperature hyperparameter.
Gradients are stopped when computing $w_i$, preventing trivial solutions and avoiding hard prompt selection.

Using the relative prompt quality $ \tilde{q}_i $, we define a group-aware regularization objective:
\begin{equation}
\mathcal{L}_{group}
=
- \sum_{i=1}^{K} \tilde{q}_i \log w_i.
\end{equation}

This regularization aligns the learned weights with relative prompt quality without allowing direct optimization of the weights.

\subsection{Prompt Consistency Regularization and Training Objective}

While quality-aware weighting adjusts prompt importance, it does not enforce prediction consistency. We therefore introduce a logit-level consistency regularization.

Let $ Z_i $ denote the predicted mask logits (before sigmoid) for prompt $ p_i $.
We select the first prompt in the group as a reference and define:
\begin{equation}
\mathcal{L}_{cons}
=
\frac{1}{K-1}
\sum_{i=2}^{K}
\big\|
Z_i - \text{stopgrad}(Z_1)
\big\|_2^2.
\end{equation}
which objective encourages consistent predictions across prompts. A stop-gradient is applied to the reference logit to avoid mutual reinforcement. Prompts are shuffled during training to prevent bias.

\paragraph{Overall Training Objective and Inference}

The final training objective for a prompt group is:
\begin{equation}
\mathcal{L}
=
\frac{1}{K}\sum_{i=1}^{K} \mathcal{L}_{seg}^{(i)}
\;+\;
\lambda \mathcal{L}_{group}
\;+\;
\beta \mathcal{L}_{cons},
\end{equation}
where $ \lambda $ and $ \beta $ control the strength of quality-aware weighting and prompt consistency enforcement, respectively.

\section{Experiments}

\subsection{Datasets and Referring Protocol}

All experiments follow the segmentation protocol of \cite{wu2025sampo}. 
Models are trained on PanNuke \cite{gamper2020pannuke} and CoNSeP \cite{graham2019hover} using only 10\% of training images to simulate data-efficient clinical scenarios. 
Evaluation is conducted on unseen datasets (CPM15, CPM17 \cite{vu2019methods}, Histology \cite{he2023transnuseg}, Kumar \cite{kumar2017dataset}, CryoNuSeg \cite{mahbod2021cryonuseg}) under strict cross-dataset generalization.

Each image--target pair is associated with a prompt group, consisting of multiple textual prompts that refer to the same segmentation target and therefore share an identical ground-truth mask.
We consider two task settings throughout the experiments: T1 (all-nuclei segmentation), where prompts refer to all nuclei in the image, and T2 (category-specific segmentation), where prompts refer to nuclei of a specific cell category.
Prompt groups are generated in a controlled and reproducible manner by varying linguistic completeness and specificity while keeping supervision unchanged.

Prompts are constructed along two orthogonal dimensions: prompt type (T1: all nuclei vs. T2: category-specific nuclei) and prompt quality, which controls the level of linguistic detail.
We define three prompt quality levels: low, medium, and high (see Table~\ref{tab:prompt_info}). 
Lower-quality prompts are intentionally short and underspecified, whereas higher-quality prompts provide richer semantic and contextual constraints.
This design enables controlled and reproducible analysis of robustness to prompt variation without introducing artificial noise.

\subsection{Experimental Setup and Baselines}

All models use AdamW ($\text{lr}=1\times10^{-4}$, batch size 1, 5 epochs) on identical data splits.
During training, prompt groups with $K$ prompts are randomly sampled per image. All prompts within the same group correspond to alternative linguistic descriptions of the \emph{same} segmentation target and share an identical ground-truth mask; thus, this setting does not introduce additional labeled data or extra semantic supervision.
In this sense, multiple prompts act as structured language-space augmentations that enforce prompt-invariant predictions, increasing the number of consistency constraints rather than the amount of supervision. All methods are evaluated using a single prompt at inference time without ensembling.
Default hyperparameters: $\tau=1.0$, $\lambda=1.0$, $\beta=0.1$.
All experiments run on a single MetaX C500 GPU (64\,GB), ensuring practical reproducibility.

\paragraph{Baselines.}
\textbf{Vision-prompt methods}: HSAM \cite{cheng2024unleashing}, SAN \cite{sun2024segment}, InstaSAM \cite{nam2024instasam}, MedSAM \cite{ma2024segment} (trained separately for T1/T2; results from prior work).
\textbf{Text-prompt SOTA}: CLIP-Seg \cite{luddecke2022clipseg}, Grounded-SAM2 \cite{ren2024groundedsam}, SAM3 \cite{carion2025sam3} (fully fine-tuned on identical schedules; reported separately for T1/T2 for fair comparison).
\textbf{Large autoregressive models}: SegZero \cite{liu2025segzero}, VisionReasoner \cite{liu2025visionreasoner} (test-only evaluation due to prohibitive computational costs).

\subsection{Quantitative Results}

\begin{figure}[t]
    \centering
    \includegraphics[width=\linewidth]{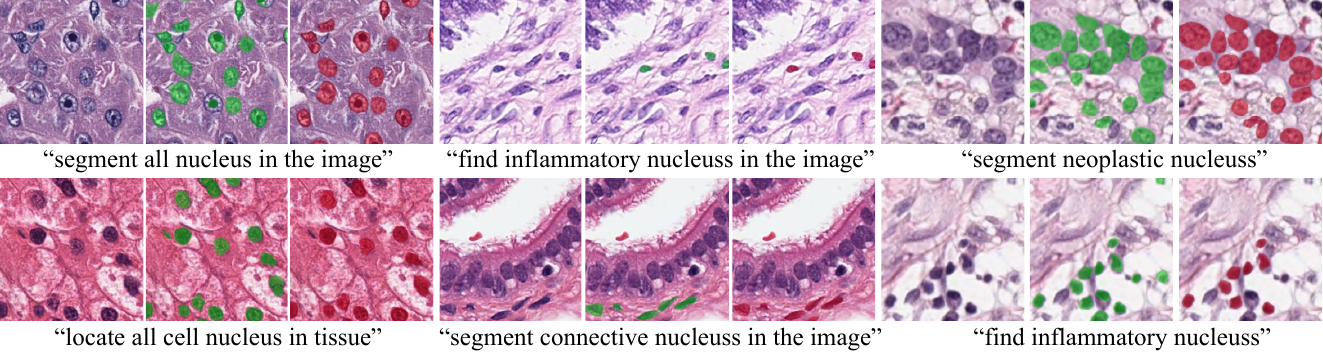}
    \caption{Qualitative comparison of segmentation results under different prompts. }
    \label{fig:vis}
\end{figure}

Table~\ref{tab:sota} reports quantitative results under visual and textual prompting. 
Under text prompts, our method performs best on both PanNuke (79.42 / 62.01) and CoNSeP (76.81 / 46.86), 
surpassing the strongest text baseline SAM3$^*$ by +0.97/+6.20 and +1.78/+3.24 Dice (T1/T2), respectively. 
Improvements are more pronounced for category-specific segmentation (T2), reflecting stronger fine-grained semantic grounding. 
Compared with prior text-driven and autoregressive models, our approach consistently achieves higher accuracy while relying solely on textual prompts at inference.
Qualitative examples are shown in Figure~\ref{fig:vis}.

\begin{table}
\centering
\caption{Segmentation performance (Dice $\uparrow$) under different methods. Visual and Text columns indicate the type of prompts used for inference. 
For text-prompt methods, results are reported in the format ``T1 / T2'', where T1 denotes all-nuclei segmentation and T2 denotes category-specific nuclei segmentation. }
\label{tab:sota}
\begin{tabular}{lcccc}
\hline
\multirow{2}{*}{Method} & \multicolumn{2}{c}{PanNuke} & \multicolumn{2}{c}{CoNSeP} \\
 & Visual & Text & Visual & Text \\
\hline
H-SAM                 & 64.90 / 40.36 & - & 18.56 / 29.76 & - \\
SAN                   & 67.77 / 40.41 & - & - & - \\
InstaSAM              & 74.67 / 21.21 & - & - & - \\
MedSAM                & 78.43 / 47.11 & - & 21.23 / 31.35 & - \\
CLIP-Seg$^*$          & - & 65.53 / 47.19 & - & 63.18 / 37.90 \\
GroundedSAM2$^*$      & - & 73.60 / 55.81 & - & 72.35 / 43.62 \\

Seg-Zero              & - & 26.01 / 12.75 & - & 22.04 / 10.57 \\
VisionReasoner        & - & 47.95 / 17.90 & - & 45.18 / 17.77 \\
SAM 3                 & - & 76.53 / 42.63 & - & 68.15 / 26.22 \\
SAM 3$^*$       & - & 78.45 / 48.98 & - & 75.03 / 40.89 \\
\hline
Ours & - & \makecell{\textbf{79.42 / 62.01} \\ \textcolor{r}{+0.97} / \textcolor{r}{+6.2}} & - & \makecell{\textbf{76.81 / 46.86} \\ \textcolor{r}{+1.78} / \textcolor{r}{+3.24}} \\
\hline
\end{tabular}
\end{table}
\vspace{0cm}

\subsection{Ablation and Sensitivity Analysis}

We conduct ablation studies to validate key design choices and analyze model sensitivity to hyperparameters.

\begin{table}[t]
\centering
\caption{Ablation and sensitivity analysis on low-quality prompts (Dice $\uparrow$). Results are reported in the format ``T1 / T2'' (all-nuclei / category-specific).}
\label{tab:ablation_all}
\begin{tabular}{ll|rc}
\hline
\multicolumn{2}{l|}{\textit{Loss design}} & \multicolumn{2}{c}{\textit{Hyperparameter sensitivity}} \\ \hline
Setting & Dice & Setting & Dice \\
\hline
Full model & \textbf{79.42} / \textbf{62.01} &  \\
w/o $\mathcal{L}_{group}, \mathcal{L}_{cons}$ & 78.45 / 48.98 & $K=2$ & 78.10 / 61.52 \\
w/o $\mathcal{L}_{cons}$ & 77.72 / 54.30                      & $K=3$ & \textbf{79.42} / \textbf{62.01} \\
$\mathcal{L}_{cons}$ (full pairwise) & 77.38 / 51.97          & $K=4$ & 79.92 / 61.24  \\
$\beta = 0.05$ & 78.50 / 61.36                                & $K=5$ & 79.11 / 60.75 \\
$\beta = 0.2$ & 78.06 / 60.86                                 & $K=6$ & 78.41 / 60.14 \\
\hline
\end{tabular}
\end{table}

Group-aware supervision is essential. Removing $\mathcal{L}_{group}$ and $\mathcal{L}_{cons}$ drops performance to 78.45 / 48.98, demonstrating that per-prompt supervision alone cannot handle prompt variability. The consistency loss further improves alignment, with performance degrading to 77.72 / 54.30 when removed.
Full pairwise consistency (77.38 / 51.97) underperforms the proposed stop-gradient formulation, suggesting that naive all-to-all alignment introduces optimization conflicts. Our stability-aware design mitigates this by using a single reference, achieving better convergence.
Performance peaks around $K=3$-$4$ and gradually declines for $K \geq 5$, indicating diminishing returns from additional prompt variations. The method remains stable across this range, with the loss weight $\beta$ showing robustness around the default value of 0.1.

\subsection{Robustness to Prompt Quality Variation}

\begin{table}[t]
\centering
\small
\caption{Overview of prompt types and quality levels used in our referring segmentation protocol. All prompts within the same group refer to the same segmentation target and share identical ground-truth masks.}
\label{tab:prompt_info}
\begin{tabular}{c|c|p{6.5cm}}
\hline
\textbf{Dimension} & \textbf{Category} & \textbf{Description} \\ \hline
\multirow{2}{*}{Prompt type}
& All
& Refers to all nuclei in the image without category distinction (e.g., ``all nuclei'', ``all cell nuclei in the tissue'') \\
& Category
& Refers to nuclei of a specific cell category (e.g., inflammatory nuclei, epithelial cell nuclei) \\ \hline
\multirow{3}{*}{Prompt quality}
& Low
& Short and underspecified prompts with limited linguistic cues; typically contain a single verb and object \\
& Medium
& Moderately descriptive prompts with paraphrasing or partial specification, including optional spatial context \\
& High
& Explicit and detailed prompts with richer verbs, spatial constraints, and occasional exclusion clauses \\ \hline
\end{tabular}
\end{table}

To analyze robustness under linguistic variation, we evaluate all methods across structured prompt quality levels (\texttt{low}, \texttt{medium}, \texttt{high}).
As shown in Table~\ref{tab:prompt_quality}, baseline methods exhibit pronounced performance degradation as prompt quality decreases.
In contrast, our method degrades more gracefully and consistently maintains higher segmentation accuracy, with the largest gains observed under low-quality prompts.

\begin{table}[!ht]
\centering
\small
\caption{Segmentation performance under different prompt quality levels. Low, Medium, and High denote prompts with increasing linguistic specificity and semantic completeness (see Table~\ref{tab:prompt_info}). Results are reported in the format ``T1 / T2'' (all-nuclei / category-specific).}
\label{tab:prompt_quality}
\begin{tabular}{c|c|c|c}
\hline
Methods & Low & Medium & High \\ \hline
SAM3      & 77.28 / 47.20 & 78.75 / 47.64 & 78.64 / 48.23 \\ \hline
Ours      & \textbf{78.75} / \textbf{62.54} & \textbf{78.77} / \textbf{62.80} & \textbf{78.89} / \textbf{61.86} \\ \hline
\end{tabular}
\end{table}

\paragraph{Zero-shot cross-domain generalization.}

\begin{table*}[!ht]
\centering
\small
\caption{Zero-shot cross-domain generalization results for \textbf{all-nuclei segmentation}. Best results are shown in \textbf{bold}. Kumar-Same and Kumar-Diff denote test sets with organ types seen and unseen during training, respectively.}
\label{tab:zeroshot}
\begin{tabular}{c|c|c|c|c|c|c}
\hline
Methods   & Histology & CPM15 & CPM17 & Kumar-same & Kumar-diff & CryoNuSeg \\ \hline
SAM2      & 46.91 & 43.10 & 42.29 & 36.29 & 40.77 & 35.78\\
SAM3      & 52.46 & 56.07 & 57.87 & 34.53 & 53.22 & 64.10\\
SAMPO     & 70.16 & 79.28 & 81.14 & 78.45 & 83.19 & 77.94\\ \hline
Ours      & \makecell{\textbf{74.61}\\\textcolor{r}{(+4.45)}} & \makecell{\textbf{79.56}\\\textcolor{r}{(+0.28)}} & \makecell{\textbf{86.10}\\\textcolor{r}{(+4.96)}} & \makecell{\textbf{80.50}\\\textcolor{r}{(+2.05)}} & \makecell{\textbf{84.78}\\\textcolor{r}{(+1.59)}} & \makecell{\textbf{77.58}\\\textcolor{g}{(-0.36)}}\\ \hline
\end{tabular}
\end{table*}

We further evaluate zero-shot generalization on multiple external datasets covering diverse tissue types, imaging modalities, and acquisition protocols.
None of the evaluated models are trained or fine-tuned on these target datasets.

SAM2/SAMPO use standard visual prompts, while SAM3 and ours rely solely on text with identical construction across datasets.
As shown in Table~\ref{tab:zeroshot}, our method achieves the best results on most benchmarks (Histology, CPM15/17, Kumar), remaining competitive with vision-prompt-based SAMPO.
Performance slightly decreases on CryoNuSeg, yet overall results demonstrate strong zero-shot transferability.

\section{Conclusion and Limitations}

We present a prompt group–aware training framework that improves robustness in text-guided medical image segmentation by modeling semantically equivalent prompts with a shared target. Experiments on nuclei segmentation show enhanced stability across diverse textual descriptions, indicating good generalizability.

Our approach adopts a fixed text encoder to enable controlled and efficient analysis of prompt variability, which may limit the modeling of highly complex semantics. Future work will explore integrating more expressive text encoders, such as large language models, and developing more advanced preference-based optimization strategies to further enhance robustness and semantic understanding.

\bibliographystyle{splncs04}
\bibliography{refs}

\end{document}